# Comprehensive Survey of Model Compression and Speed up for Vision Transformers

**Feiyang Chen [1], Ziqian Luo [2], Lisang Zhou[3], Xueting Pan[4], Ying Jiang [5]**

[1] *Coupang, Mountain View, 94043, CA, United States, Email: feiyang.chen001@gmail.com*
[2]*Oracle, Seattle, 98101, WA, United States, Email: luoziqian98@gmail.com*
[3]*Bazaarvoice Inc., Austin, 78759, TX, United States, Email: lzhou@berkeley.edu*
[4] *Oracle, Seattle, 98101, WA, United States, Email: xtpan8800@gmail.com*
[5] *Carnegie Mellon University, Pittsburgh, 15213, PA, United States, Email: yingj2@alumni.cmu.edu*

## Abstract

Vision Transformers (ViT) have marked a paradigm shift in computer vision, outperforming state-of-the-art models across diverse tasks. However, their practical deployment is hampered by high computational and memory demands. This study addresses the challenge by evaluating four primary model compression techniques: quantization, low-rank approximation, knowledge distillation, and pruning. We methodically analyze and compare the efficacy of these techniques and their combinations in optimizing ViTs for resource-constrained environments. Our comprehensive experimental evaluation demonstrates that these methods facilitate a balanced compromise between model accuracy and computational efficiency, paving the way for wider application in edge computing devices.

## 1 Introduction

Transformers, as introduced by Vaswani et al. [1], have revolutionized machine learning with their high capacity for modeling data, scalability, and exceptional aptitude for capturing long-range dependencies. Originating in the natural language processing (NLP) sphere with successes in machine translation and text summarization [2], Transformers now extend their utility to computer vision tasks such as image classification [3] and object detection [4]. More recently, their prowess has been harnessed in the field of multimodal sentiment analysis [5], where they integrate and interpret diverse data streams—textual, visual, and acoustic [6, 7]—to assess emotional responses [8]. The application of Transformers has also expanded into the healthcare domain [9], facilitating advancements in areas such as disease detection [10], medical imaging analysis [11, 12], and the processing of electronic health records. These technologies are pivotal in developing predictive models for patient outcomes, enhancing diagnostic accuracy, and personalizing treatment strategies, showcasing the potential of Transformers to impact patient care significantly. Unlike previous research in NLP that grapples with the quadratic complexity of softmax-attention in lengthy sequences as noted by Child et al. [13], Vision Transformers (ViTs) manage shorter, fixed-length input sequences. As a result, softmax-attention represents a reduced share of the total floating-point operations (FLOPs) in ViTs, presenting a distinct set of optimization challenges and underscoring the need for specialized model compression strategies.

The Vision Transformer (ViT) [14], an innovative architecture derived from NLP methodologies, has demonstrated that transforming image patches into token sequences processed by transformer blocks can achieve, and sometimes surpass, the accuracy of traditional CNN architectures. This breakthrough has catalyzed a paradigm shift in image processing. However, the advanced performance of ViT is accompanied by a substantially large parameter set, numbering in the hundreds of millions, which results in considerable memory and computational overhead during inference. Such demands render ViTs less viable for devices with limited resources or power constraints. Consequently, the development and refinement of model compression techniques for ViTs have become critical for their adoption in industrial applications, where efficiency and resource management are paramount.

In contrast to well-explored deep learning models, the model compression landscape for Vision Transformers remains relatively untapped. This paper presents a systematic exploration of model compression strategies for ViTs, focusing on quantization, low-rank approximation, knowledge distillation, and pruning. Through a thorough comparative analysis, we assess the individual and collective impact of these techniques on model efficiency. Our investigation also delves into the potential synergistic effects of combining different methods, with the goal of enhancing performance. The extensive experimental findings confirm that our proposed approaches facilitate a favorable balance between maintaining accuracy and improving computational efficiency, which is essential for practical deployments.

## 2 Related work

### 2.1 Quantization

Quantization has emerged as a cornerstone technique for promoting efficient inference in neural networks. This process involves converting a network into a low-bit representation, thereby reducing both computational demands and memory footprint, with minimal impact on model performance. A critical aspect of this technique is establishing an appropriate clipping range for the weights. Krishnamoorthi [15] suggests determining this range by evaluating all weights within a layer's convolutional filters, whereas Shen et al. [16] adopt a groupwise quantization approach for Transformers. To mitigate the potential loss in accuracy that quantization may introduce, Quantization-Aware Training (QAT) has been proposed. QAT involves conducting the standard forward and backward passes using the floating-point representation of the quantized model, followed by the re-quantization of model parameters after each gradient update, thus preserving accuracy while leveraging the benefits of quantization.

### 2.2 Low-rank Approximation

The Vision Transformer (ViT) capitalizes on the self-attention mechanism, which inherently entails a quadratic computational complexity, a significant challenge for scalability. Chen et al. [17] highlight that the attention matrices within ViTs inherently possess low-rank properties [18], presenting an opportunity for complexity reduction [19]. Leveraging low-rank matrix approximation on these attention matrices emerges as a promising approach to mitigating computational costs. A variety of methodologies have been developed for this purpose, including Nyström-based methods [20, 21], Performer [22], and Linformer [23], each with its own unique implementation and suitability for integration with pre-trained ViT models during fine-tuning and validation phases. Moreover, the combination of low-rank approximation with sparse attention mechanisms, as suggested by Chen et al. [17], has been shown to yield even more refined approximations, enhancing the efficiency and effectiveness of ViTs.

### 2.3 Knowledge Distillation

Knowledge distillation is a refined model compression technique wherein a compact 'student' model is trained to emulate a more complex 'teacher' model by utilizing the teacher's soft labels. These soft labels are recognized for their rich informational content, often leading to superior performance in the student model as compared to training with hard labels [24]. The efficacy of soft labels in enhancing student learning has been corroborated by Yuan et al. [25] and Wei et al. [26]. A novel advancement in this domain is the introduction of a distillation token in Vision Transformers [14] by Touvron et al. [27]. This token, analogous to the class token but dedicated to capturing the teacher's predictions, engages through the self-attention mechanism, optimizing the distillation process. Such bespoke approaches have demonstrated considerable gains over conventional distillation techniques, underscoring the potential for transformer-specific optimization strategies.

### 2.4 Pruning

Pruning represents a widely endorsed method to streamline the architecture of Vision Transformers by reducing their dimensional complexity [28]. Central to this technique is the assignment of an importance score to each model dimension, allowing for the selective elimination of dimensions deemed less critical, based on their scores. This targeted reduction aims to maintain a robust pruning ratio while preserving the model's accuracy. The strategy of dimensional redistribution, as proposed by Yang et al. [29], may be integrated into the pruning process, further refining the model's performance. Intriguingly, studies have shown that a model, post-pruning, can occasionally surpass the original in performance, indicating the potential of pruning to not only simplify but also to enhance the functionality of the model [30].

## 3 Methodology

### 3.1 Quantization

#### 3.1.1 Basic Concept

The overarching objective of quantization is to reduce the precision of model parameters ($\vartheta$) and intermediate activation maps to a lower precision format, such as 8-bit integers, while minimizing the impact on the model's generalization performance. The initial step in this process involves defining a quantization function capable of mapping weights and activations to a discrete set of values. A commonly utilized function for this purpose is delineated as follows:

$$Q(r) = \text{Int}(r/S) - Z, \tag{1}$$

where $Q$ represents the quantization mapping function, $r$ denotes a real-valued input (e.g., weights, activation), $S$ is a scaling factor, and $Z$ is an integer zero point. This mechanism, known as *uniform quantization*, ensures the equidistant spacing of resultant values. It's noteworthy that alternative *non-uniform quantization* strategies exist. Moreover, the original real value $r$ can be approximated from its quantized counterpart $Q(r)$ through a process known as *dequantization*:

$$\tilde{r} = S(Q(r) + Z), \tag{2}$$

where the approximation $\tilde{r}$ may differ from $r$ due to rounding errors inherent in quantization.

A critical aspect of quantization is determining the optimal scaling factor $S$, which effectively partitions real values $r$ into discrete segments:

$$S = \frac{\beta - \alpha}{2^b - 1}, \tag{3}$$

with [$\alpha$, $\beta$] representing the clipping range and $b$ denoting the bit width of quantization. The selection of the clipping range [$\alpha$, $\beta$], a process termed as *calibration*, is pivotal. A straightforward method involves employing the minimum and maximum of the inputs as the clipping range, i.e., $\alpha = r_{\min}$ and $\beta = r_{\max}$, corresponding to an *asymmetric quantization* scheme where $-\alpha \neq \beta$. Alternatively, a *symmetric quantization* approach, where $-\alpha = \beta = \max(|r_{\max}|, |r_{\min}|)$, can be employed. In such cases, the quantization function in Eq. 1 can be simplified by setting $Z = 0$.

#### 3.1.2 Post Training Quantization

Post Training Quantization (PTQ) streamlines the quantization process by adjusting weights directly, without necessitating further fine-tuning. This efficiency, however, may lead to notable accuracy declines due to the inherent precision loss of quantization. Liu et al. [31] observed substantial accuracy reductions when applying quantization to LayerNorm and Softmax layers within Transformer architectures. Lin et al. [32] attributed these discrepancies to the polarized distribution of activation values in LayerNorm layers and attention map values. Specifically, significant inter-channel variability within LayerNorm layer inputs (as illustrated on the left side of Figure 1) induces considerable quantization errors when employing layer-wise quantization approaches. Moreover, a predominance of small-value distributions in attention maps—with only sparse outliers approaching a value of 1—further exacerbates performance declines under uniform quantization strategies. Addressing these challenges, Lin et al. [32] introduced a novel quantization approach employing Powers-of-Two Scale for LayerNorm and Log-Int-Softmax for Softmax layers, aiming to mitigate the adverse effects of traditional quantization methods.

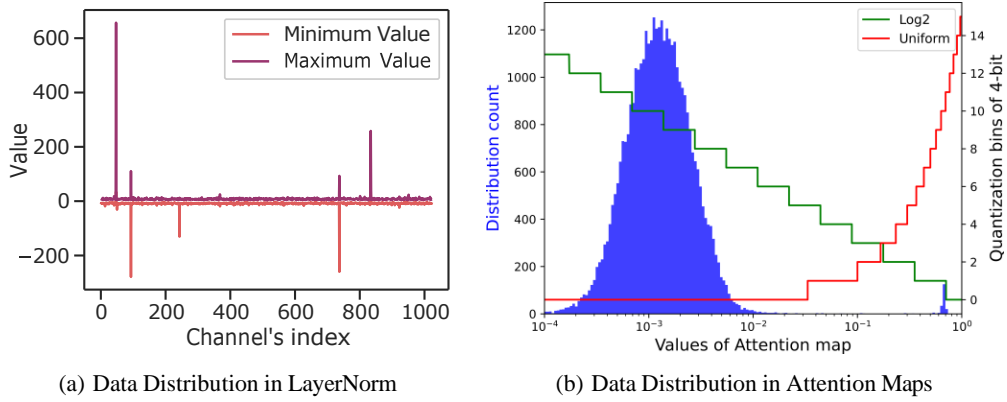

(a) Data Distribution in LayerNorm
(b) Data Distribution in Attention Maps

Figure 1: **Left [32]**: Channel-wise minimum and maximum values of the last LayerNorm inputs in full precision Swin-B. **Right [32]**: Distribution of the attention map values from the first layer of ViT-L, and visualizing the quantized bins using uniform or Log2 quantization with 4-bit.

### 3.1.3 Quantization Aware Training

Applying quantization directly to a fully trained model can inadvertently perturb model parameters, leading to notable performance declines. An effective strategy to circumvent this issue involves re-training the model with quantized parameters, thereby guiding the model towards a more favorable loss landscape. Quantization Aware Training (QAT) stands out as a prominent technique for this purpose. In QAT, the standard forward and backward processes are executed on a model represented in floating-point, yet parameters are re-quantized following each gradient update, ensuring the model adapts to quantization-induced changes.

Learned Step Size Quantization (LSQ) [33], an advancement in this field, refines the quantizer configuration process and has set new benchmarks in quantization performance by optimizing the quantization intervals. Similarly, DIFFQ [34] introduces a differentiable model compression technique that eschews the need for gradient approximation methods such as the Straight Through Estimator (STE). By employing pseudo quantization noise, DIFFQ achieves an approximation of the quantization process during training that is fully differentiable, thereby facilitating more nuanced adjustments to both the weights and quantization bit-depth.

## 3.2 Knowledge Distillation

Knowledge distillation techniques, such as soft and hard distillation, facilitate the transfer of knowledge from a complex 'teacher' model to a simpler 'student' model. Soft distillation focuses on minimizing the Kullback-Leibler (KL) divergence between the softened logits (or outputs) of both the teacher and student models. This is formally captured by the distillation objective:

$$L_{\text{global}} = (1 - \lambda) L_{\text{CE}}(\psi(Z_s), y) + \lambda \tau^2 \text{KL}\left(\psi\left(\frac{Z_s}{\tau}\right), \psi\left(\frac{Z_t}{\tau}\right)\right), \quad (4)$$

where $L_{\text{CE}}$ denotes the cross-entropy loss, $\psi$ represents the softmax function, $Z_t$ and $Z_s$ are the logits from the teacher and student models, respectively, $\tau$ is the temperature parameter enhancing softness of distributions, and $\lambda$ balances the contributions of the KL divergence and the cross-entropy loss. Conversely, hard distillation uses the teacher's predictions as definitive labels for training the student, simplifying the process by directly comparing the student's predictions against these labels:

$$L_{\text{global}}^{\text{hardDistill}} = \frac{1}{2} L_{\text{CE}}(\psi(Z_s), y) + \frac{1}{2} L_{\text{CE}}(\psi(Z_s), y_t), \quad (5)$$

where $y_t = \text{argmax}_c Z_t(c)$ represents the hard label decision by the teacher model.

The DeiT [27] method introduces a novel approach specific to Transformers, incorporating a 'distillation token' into the architecture that functions analogously to the class token but focuses on mimicking the teacher's predictions. This mechanism allows for a direct interaction between the distillation token and other components through the self-attention layers, demonstrating superior performance in distillation. Our experimental setup involves applying the DeiT framework for knowledge distillation on the CIFAR dataset, adjusting for computational resource constraints.

### 3.3 Pruning

Pruning in Vision Transformers focuses primarily on reducing the model's complexity by decreasing the number of parameters, specifically by adjusting the dimensions of weight kernels between hidden layers. This objective can be formalized as:

$$\min \alpha, \beta \quad \text{s.t.} \sum_k loss(I_\beta^{(k)} W_{\alpha,\beta}^{(k)} I_\alpha^{(k+1)}) - loss(I_b^{(k)} W_{a,b}^{(k)} I_a^{(k+1)}) < \delta \tag{6}$$

where a, b represent the original dimensions of W (k), and α, β are the reduced dimensions post- pruning. The goal is to ensure the incremental loss incurred from this reduction remains below a predefined threshold δ, preserving the integrity of the model for subsequent tasks. Determining which dimensions to prune involves the use of importance scores, a concept learned either during pre-training or fine-tuning. Zhu et al. [28] and Yang et al. [29] derive these scores from the gradient magnitude of each weight, proposing the integration of a "soft gate" layer post-pruning which hardens to zero-out less critical dimensions during inference:

$$S_B(W) = \left(\sum_{b \in B} \frac{\partial s}{\partial \omega_b} \omega_b\right)^2 \tag{7}$$

Alternatively, Yu et al. [30] employ KL divergence to calculate importance scores, focusing on the divergence between model performances with and without specific modules across a dataset Ω. This method facilitates both within-layer and across-module pruning:

$$S_B(W) = \sum_{i \in \Omega} D_{KL}(p_i || q_i) \tag{8}$$

where qi corresponds to the loss with the full model, and pi to the loss sans the pruned weights. Recent innovations have introduced even more nuanced importance scoring systems. Tang et al. [35] devised a score reflecting the theoretical impact of each patch on the final error, enhancing patch slimming efficiency. Rao et al. [36] combined local and global features for a more holistic assessment of token significance. Similarly, Yi et al. [37] synthesized various scores into a unified loss function, further refining the pruning process.

### 3.4 Low-rank Approximation

In Vision Transformers (ViTs), as shown in Figure 2, each self-attention block initially projects an input sequence $X$ using weights $W_Q$, $W_K$, and $W_V$ to obtain feature representations $Q = W_Q X$, $K = W_K X$, and $V = W_V X$. The self-attention mechanism, computed as $soft(QK^T/\sqrt{d_q})V$, introduces computational and spatial complexities of $O(n^2)$, where $n$ is the sequence length [21].

Given the formal proof of self-attention's low-rank nature [23], leveraging this property for low-rank approximation emerges as a strategic choice to enhance computational efficiency. Such approxi- mations aim to preserve accuracy while significantly reducing both time and space complexity to approximately $O(n)$, even when integrating with pre-existing or newly trained models [21, 20, 17].

Notably, low-rank approximation does not inherently reduce the model size, as the original weights WQ, WK, WV are retained. However, it does offer substantial reductions in computational time and memory usage, particularly during fine-tuning or validation phases for pre-trained models. This is because the approximation calculations are performed subsequent to input reception. Various methodologies for low-rank approximation exist, including Nyström-based approaches like Nyströmformer [21] and SOFT [20], which linearize self-attention through the Nyström method. Alternative linearization techniques, such as Linformer [23] and Performer [22], along with strategies integrating both low-rank and sparse attention mechanisms [17], further enhance approximation accuracy.

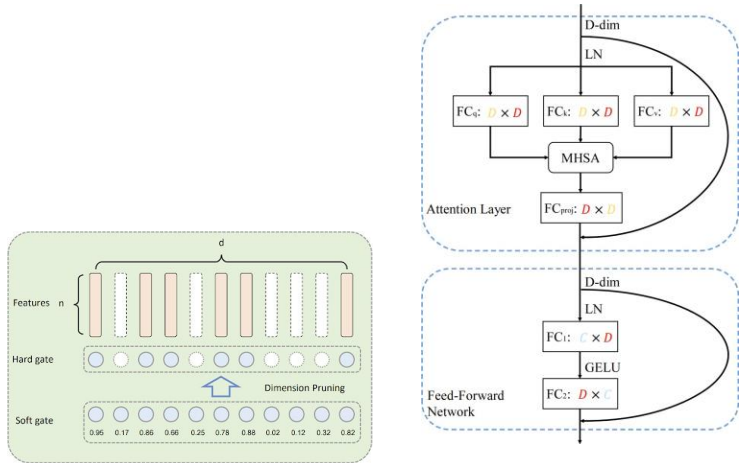

(a) Soft-gate based model proposed by Zhu et. al. [28]  (b) Skip-dimension model proposed by Yu et. al. [30]

Figure 2: Different pruning models

Our experimental focus is on the Nyströmformer-based ViT, adapting the softmax attention matrix calculation to utilize the Nyström method across all self-attention blocks. This allows for the use of pre-trained Vanilla ViT weights, facilitating direct comparisons in performance. The Nyströmformer technique employs landmark points for an efficient approximation, circumventing the need for full QKT calculations. We evaluate the efficacy of this approach with landmark counts (m) set at 24, 32, and 64, assessing its impact on model performance.

## 4 Experiments

This section delineates a thorough comparative analysis of various model compression techniques applied to Vision Transformers, including quantization, knowledge distillation, pruning, and low-rank approximation. Additionally, we investigate the synergistic potential of combining these methods to ascertain enhancements in performance metrics.

### 4.1 Experimental Settings

The experimental framework is established on a Tesla V100-SXM2 16GB GPU, with PyTorch serving as the primary platform for code implementation[1]. The scope of our dataset utilization is confined to CIFAR-10 and CIFAR-100, attributed to computational resource constraints. Our primary metrics of interest include model size and inference speed, acknowledging the inherent trade-off between accuracy and these efficiency parameters. An optimal compression technique is thus characterized by minimal impact on accuracy coupled with substantial reductions in model size and enhancements in inference speed. Results of the comparative analysis across CIFAR-10 and CIFAR-100 datasets are systematically presented in Table 1 and Table 2.

### 4.2 Comparison of Different Model Compression Methods

In assessing the impact of model compression on **Model Size**, we find that quantization and pruning strategies offer substantial size reductions with minimal accuracy compromise. Notably, quantization techniques, particularly Dynamic Quantization[2], have demonstrated superior efficacy, reducing model size to 25

Contrarily, weight pruning, particularly with simplistic importance scoring, does not facilitate an optimal balance between model size and accuracy. A pruning rate of 0.1 (indicating 10% of parameters pruned) led to a significant accuracy reduction in both CIFAR-10 and CIFAR-100 datasets compared to the unpruned ViT. Further investigation, as depicted in figure 3, reveals that a majority of parameters are deemed critically important (scores above 0.99), suggesting inherent limitations in simple form importance scoring for weight pruning. Enhancements could stem from the integration of more sophisticated importance scores [30] or adopting strategies like input patch reduction or slimming, as opposed to direct weight pruning [36, 35].

---

[1] It is pertinent to note that PyTorch's current support for quantization is limited to CPU-based operations, necessitating CPU-based inference speed tests for certain methodologies.

[2] Dynamic Quantization was implemented using the PyTorch Quantization API: https://pytorch.org/tutorials/advanced/dynamic_quantization_tutorial.html

Table 1: Evaluation results on CIFAR-10. The Speed values are iterations per second

| Model | Method | Accuracy | GPU Speed | CPU Speed | Size(MB) |
|---|---|---|---|---|---|
| Vanilla ViT [14] | - | 98.94 | 4.48 | 0.050 | 327 |
| Dynamic Quantization | Quantization (PTQ) | 98.73 | - | 0.062 | 84 |
| FQ-ViT [32] | Quantization (PTQ) | 97.31 | - | - | - |
| DIFFQ with LSQ [33] | Quantization (QAT) | 93.37 | 2.10 | - | 41 |
| DIFFQ with diffq [34] | Quantization (QAT) | 60.29 | 12.20 | - | 2 |
| DeiT base [27] | Knowledge Distillation | 98.47 | 7.04 | 0.096 | 327 |
| DeiT tiny [27] | Knowledge Distillation | 95.43 | 16.78 | - | 21 |
| ViT-Pruning(r=0.1) [28] | Pruning | 88.36 | 4.86 | - | 301 |
| ViT-Pruning(r=0.2) [28] | Pruning | 80.56 | 5.54 | - | 254 |
| ViT-Nyströmformer(m=24) [21] | Low-rank Approximation | 65.91 | 4.67 | - | 327 |
| ViT-Nyströmformer(m=32) [21] | Low-rank Approximation | 75.94 | 4.57 | - | 327 |
| ViT-Nyströmformer(m=64) [21] | Low-rank Approximation | 91.70 | 4.38 | - | 327 |
| DeiT base + Dynamic Quantization | Knowledge Distillation + PTQ | 96.75 | - | 0.117 | 84 |

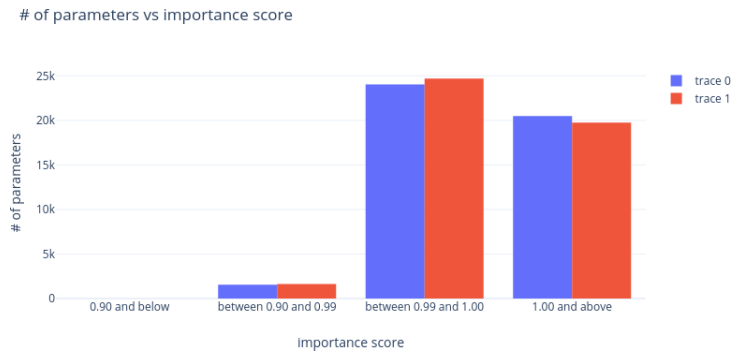

Figure 3: Number of parameters vs importance score. **Blue**: CIFAR-10. **Red**: CIFAR-100.

For **Inference Speed**, a spectrum of enhancements is observed across different model compression strategies, with methods centered around knowledge distillation particularly standing out for their efficiency gains. Notably, the DeiT base model, despite not undergoing significant size reduction, achieves an inference speed nearly double that of the standard Vision Transformer (ViT), all while preserving accuracy to a remarkable degree. An intriguing case is observed with the DeiT tiny configuration on the CIFAR-10 dataset, where it attains 95.43% accuracy—a figure closely aligned with the Vanilla ViT—yet delivers a quadruple increase in speed and is compressed to merely 6% of the original model's size.

Furthermore, the application of Nyströmformer-based techniques to ViT illustrates a nuanced balance between accuracy and speed, particularly influenced by the selection of the number of landmarks ($m$). Opting for a larger $m$ value enhances the precision of approximations at the expense of processing velocity. Additionally, Dynamic Quantization contributes to inference speed improvements in the range of 10-20% on CPU platforms, underscoring the practical benefits of model compression beyond just reductions in size.

Table 2: Evaluation results on CIFAR-100. The Speed values are iterations per second.

| Model | Method | Accuracy | GPU Speed | CPU Speed | Size(MB) |
|---|---|---|---|---|---|
| Vanilla ViT [14] | - | 92.87 | 4.34 | 0.093 | 327 |
| Dynamic Quantization | Quantization (PTQ) | 90.87 | - | 0.122 | 84 |
| FQ-ViT [32] | Quantization (PTQ) | 84.87 | - | - | - |
| DIFFQ with LSQ [33] | Quantization (QAT) | 76.08 | 2.10 | - | 41 |
| DIFFQ with diffq [34] | Quantization (QAT) | 41.02 | 12.00 | - | 2 |
| DeiT base [27] | Knowledge Distillation | 87.35 | 6.97 | 0.149 | 327 |
| DeiT tiny [27] | Knowledge Distillation | 75.90 | 16.16 | - | 21 |
| ViT-Pruning(r=0.1) [28] | Pruning | 74.46 | 4.69 | - | 302 |
| ViT-Pruning(r=0.2) [28] | Pruning | 64.27 | 5.19 | - | 272 |
| ViT-Nyströmformer(m=24) [21] | Low-rank Approximation | 38.51 | 4.77 | - | 327 |
| ViT-Nyströmformer(m=32) [21] | Low-rank Approximation | 50.31 | 4.65 | - | 327 |
| ViT-Nyströmformer(m=64) [21] | Low-rank Approximation | 74.01 | 4.46 | - | 327 |
| DeiT base + Dynamic Quantization | Knowledge Distillation + PTQ | 82.61 | - | 0.196 | 84 |

### 4.3 Exploration of Mixed Methods

The examination of individual model compression techniques suggests that a hybrid approach, leveraging the strengths of both quantization and knowledge distillation, warrants further investigation. Particularly, when a slight decrease in accuracy is acceptable, such a combined strategy appears promising for optimizing both model compactness and processing efficiency. As demonstrated in Tables 1 and 2, employing a composite method—integrating the DeiT base model with Dynamic Quantization—significantly enhances inference speed, achieving more than a twofold increase, while concurrently reducing the model's size to one-fourth of its original dimensions. This is achieved with a manageable trade-off in accuracy, highlighting the potential of mixed approaches in striking a balanced compromise between speed, size, and performance.

## 5 Conclusion

This study has been dedicated to an empirical investigation of model compression techniques aimed at enhancing the efficiency and deployment viability of Vision Transformers (ViTs). We meticulously examined four predominant compression methods—quantization, low-rank approximation, knowledge distillation, and pruning—complemented by a review of cutting-edge research in the field. Through comparative analyses conducted on the CIFAR-10 and CIFAR-100 datasets, our findings underscore the efficacy of post-training quantization and knowledge distillation as standout strategies. These methods not only significantly reduce model size but also expedite inference times, all while maintaining acceptable levels of performance degradation. Further exploration into the synergistic potential of combining quantization and knowledge distillation has revealed a compelling avenue for optimization. Particularly evident within the CIFAR-10 dataset, this hybrid approach markedly accelerated inference speeds—surpassing baseline speeds by more than a factor of two—while concurrently diminishing model size to merely a quarter of its initial footprint. The insights garnered from this comprehensive examination advocate for a holistic, multi-faceted approach to model compression. Integrating diverse compression methodologies holds substantial promise for refining the operational efficiency of Vision Transformers, heralding a robust direction for future research in this domain. Moreover, this paper can be further investigated in the fields of transportation engineering [38-42], machine learning [43-46], biological engineering [47-51], etc.

# References

[1] A. Vaswani et al., "Attention is All you Need," in Advances in Neural Information Processing Systems, Curran Associates, Inc., 2017.

[2] Z. Luo, "Knowledge-guided Aspect-based Summarization," in 2023 International Conference on Communications, Computing and Artificial Intelligence (CCCAI), Jun. 2023, pp. 17–22. doi: 10.1109/CCCAI59026.2023.00012.

[3] M. FCNCH and H. Hu, "Assessing four neural networks on handwritten digit recognition dataset (MNIST)," CoRR, vol. abs/1811.08278, 2018.

[4] Z. Luo, X. Zeng, Z. Bao, and M. Xu, "Deep learning-based strategy for macromolecules classification with imbalanced data from cellular electron cryotomography," in 2019 International Joint Conference on Neural Networks (IJCNN), IEEE, 2019, pp. 1–8.

[5] F. Chen and Z. Luo, "Sentiment Analysis using Deep Robust Complementary Fusion of Multi-Features and Multi-Modalities," CoRR, 2019.

[6] Z. Luo, H. Xu, and F. Chen, "Utterance-based audio sentiment analysis learned by a parallel combination of cnn and lstm," arXiv preprint arXiv:1811.08065, 2018.

[7] Z. Luo, H. Xu, and F. Chen, "Audio Sentiment Analysis by Heterogeneous Signal Features Learned from Utterance-Based Parallel Neural Network.," in AffCon@ AAAI, Shanghai, China, 2019, pp. 80–87.

[8] F. Chen and Z. Luo, "Learning Robust Heterogeneous Signal Features from Parallel Neural Network for Audio Sentiment Analysis." arXiv, Jul. 31, 2019.

[9] J. Zhang et al., "Predicting unseen antibodies' neutralizability via adaptive graph neural networks," Nature Machine Intelligence, vol. 4, no. 11, pp. 964–976, 2022.

[10] Y. Wu, M. Gao, M. Zeng, J. Zhang, and M. Li, "BridgeDPI: a novel graph neural network for predicting drug–protein interactions," Bioinformatics, vol. 38, no. 9, pp. 2571–2578, 2022.

[11] F. Chen, Y. Jiang, X. Zeng, J. Zhang, X. Gao, and M. Xu, "PUB-SalNet: A pre-trained unsupervised self-aware backpropagation network for biomedical salient segmentation," Algorithms, vol. 13, no. 5, p. 126, 2020.

[12] S. Liu et al., "A unified framework for packing deformable and non-deformable subcellular structures in crowded cryo-electron tomogram simulation," BMC Bioinformatics, vol. 21, no. 1, p. 399, Dec. 2020, doi: 10.1186/s12859-020-03660-w.

[13] R. Child, S. Gray, A. Radford, and I. Sutskever, "Generating long sequences with sparse transformers. arXiv 2019," arXiv preprint arXiv:1904.10509, 2019.

[14] A. Dosovitskiy et al., "An Image is Worth 16x16 Words: Transformers for Image Recognition at Scale." arXiv, Jun. 03, 2021.

[15] R. Krishnamoorthi, "Quantizing deep convolutional networks for efficient inference: A whitepaper. arXiv 2018," arXiv preprint arXiv:1806.08342, 1806.

[16] S. Shen et al., "Q-bert: Hessian based ultra low precision quantization of bert," in Proceedings of the AAAI Conference on Artificial Intelligence, 2020, pp. 8815–8821.

[17] B. Chen, T. Dao, E. Winsor, Z. Song, A. Rudra, and C. Ré, "Scatterbrain: Unifying Sparse and Low-rank Attention Approximation." arXiv, Oct. 28, 2021.

[18] F. Chen, N. Chen, H. Mao, and H. Hu, "The Application of Bipartite Matching in Assignment Problem. arXiv 2019," arXiv preprint arXiv:1902.00256.

[19] F. Chen, N. Chen, H. Mao, and H. Hu, "An efficient sorting algorithm-Ultimate Heapsort (UHS). 2019."

[20] J. Lu et al., "Soft: Softmax-free transformer with linear complexity," Advances in Neural Information Processing Systems, vol. 34, pp. 21297–21309, 2021.

[21] X. Yunyang et al., "Nyströmformer: A Nystöm-based Algorithm for Approximating Self-Attention," AAAI, 2021.

[22] K. Choromanski et al., "Rethinking Attention with Performers." arXiv, Nov. 19, 2022. doi: 10.48550/arXiv.2009.14794.

[23] S. Wang, B. Z. Li, M. Khabsa, H. Fang, and H. Ma, "Linformer: Self-Attention with Linear Complexity." arXiv, Jun. 14, 2020.

[24] G. Hinton, O. Vinyals, and J. Dean, "Distilling the Knowledge in a Neural Network." arXiv, Mar. 09, 2015.

[25] L. Yuan et al., "Tokens-to-token vit: Training vision transformers from scratch on imagenet," in Proceedings of the IEEE/CVF international conference on computer vision, 2021, pp. 558–567.

[26] L. Wei, A. Xiao, L. Xie, X. Zhang, X. Chen, and Q. Tian, "Circumventing Outliers of AutoAugment with Knowledge Distillation," in Computer Vision – ECCV 2020, vol. 12348, A. Vedaldi, H. Bischof, T. Brox, and J.-M. Frahm, Eds., in Lecture Notes in Computer Science, vol. 12348. , Cham: Springer International Publishing, 2020, pp. 608–625. doi: 10.1007/978-3-030-58580-8_36.

[27] H. Touvron, M. Cord, M. Douze, F. Massa, A. Sablayrolles, and H. Jegou, "Training data-efficient image transformers & distillation through attention," in Proceedings of the 38th International Conference on Machine Learning, PMLR, Jul. 2021, pp. 10347–10357.

[28] M. Zhu, Y. Tang, and K. Han, "Vision Transformer Pruning." arXiv, Aug. 14, 2021. doi: 10.48550/arXiv.2104.08500.

[29] S. Yu et al., "Unified Visual Transformer Compression." arXiv, Mar. 15, 2022. doi: 10.48550/arXiv.2203.08243.

[30] H. Yang, H. Yin, P. Molchanov, H. Li, and J. Kautz, "Nvit: Vision transformer compression and parameter redistribution," 2021.

[31] H. Yu and J. Wu, "A unified pruning framework for vision transformers," Sci. China Inf. Sci., vol. 66, no. 7, p. 179101, Jul. 2023, doi: 10.1007/s11432-022-3646-6.

[32] Z. Liu, Y. Wang, K. Han, W. Zhang, S. Ma, and W. Gao, "Post-training quantization for vision transformer," Advances in Neural Information Processing Systems, vol. 34, pp. 28092–28103, 2021.

[33] Y. Lin, T. Zhang, P. Sun, Z. Li, and S. Zhou, "Fq-vit: Fully quantized vision transformer without retraining," arXiv preprint arXiv:2111.13824, 2021.

[34] S. K. Esser, J. L. McKinstry, D. Bablani, R. Appuswamy, and D. S. Modha, "Learned Step Size Quantization." arXiv, May 06, 2020. doi: 10.48550/arXiv.1902.08153.

[35] A. Défossez, Y. Adi, and G. Synnaeve, "Differentiable Model Compression via Pseudo Quantization Noise." arXiv, Oct. 17, 2022.

[36] Y. Tang et al., "Patch slimming for efficient vision transformers," in Proceedings of the IEEE/CVF Conference on Computer Vision and Pattern Recognition, 2022, pp. 12165–12174.

[37] Y. Rao, W. Zhao, B. Liu, J. Lu, J. Zhou, and C.-J. Hsieh, "Dynamicvit: Efficient vision transformers with dynamic token sparsification," Advances in neural information

processing systems, vol. 34, pp. 13937–13949, 2021.

[38] Y. Zhao, W. Dai, Z. Wang, and A. E. Ragab, "Application of computer simulation to model transient vibration responses of GPLs reinforced doubly curved concrete panel under instantaneous heating," Materials Today Communications, vol. 38, p. 107949, Mar. 2024, doi: 10.1016/j.mtcomm.2023.107949.

[39] W. Dai, M. Fatahizadeh, H. G. Touchaei, H. Moayedi, and L. K. Foong, "Application of six neural network-based solutions on bearing capacity of shallow footing on double-layer soils," Steel and Composite Structures, vol. 49, no. 2, pp. 231–244, 2023, doi: 10.12989/scs.2023.49.2.231.

[40] W. Dai, "Safety Evaluation of Traffic System with Historical Data Based on Markov Process and Deep-Reinforcement Learning," Journal of Computational Methods in Engineering Applications, pp. 1–14, Oct. 2021.

[41] W. Dai, "Design of Traffic Improvement Plan for Line 1 Baijiahu Station of Nanjing Metro," Innovations in Applied Engineering and Technology, Dec. 2023, doi: 10.58195/iaet.v2i1.133.

[42] W. Dai, "Evaluation and Improvement of Carrying Capacity of a Traffic System," Innovations in Applied Engineering and Technology, pp. 1–9, Nov. 2022, doi: 10.58195/iaet.v1i1.001.

[43] H. Wang, Y. Zhou, E. Perez, and F. Roemer, "Jointly Learning Selection Matrices For Transmitters, Receivers And Fourier Coefficients In Multichannel Imaging." arXiv, Feb. 29, 2024. Accessed: Mar. 23, 2024.

[44] L. Zhou, Z. Luo, and X. Pan, "Machine learning-based system reliability analysis with Gaussian Process Regression." arXiv, Mar. 17, 2024.

[45] M. Li, Y. Zhou, G. Jiang, T. Deng, Y. Wang, and H. Wang, "DDN-SLAM: Real-time Dense Dynamic Neural Implicit SLAM." arXiv, Mar. 08, 2024. Accessed: Mar. 23, 2024.

[46] Y. Zhou et al., "Semantic Wireframe Detection," 2023, Accessed: Mar. 23, 2024.

[47] G. Tao et al., "Surf4 (Surfeit Locus Protein 4) Deficiency Reduces Intestinal Lipid Absorption and Secretion and Decreases Metabolism in Mice," ATVB, vol. 43, no. 4, pp. 562–580, Apr. 2023, doi: 10.1161/ATVBAHA.123.318980.

[48] Y. Shen, H.-M. Gu, S. Qin, and D.-W. Zhang, "Surf4, cargo trafficking, lipid metabolism, and therapeutic implications," Journal of Molecular Cell Biology, vol. 14, no. 9, p. mjac063, 2022.

[49] M. Wang et al., "Identification of amino acid residues in the MT-loop of MT1-MMP critical for its ability to cleave low-density lipoprotein receptor," Frontiers in Cardiovascular Medicine, vol. 9, p. 917238, 2022.

[50] Y. Shen, H. Gu, L. Zhai, B. Wang, S. Qin, and D. Zhang, "The role of hepatic Surf4 in lipoprotein metabolism and the development of atherosclerosis in apoE-/- mice," Biochimica et Biophysica Acta (BBA)-Molecular and Cell Biology of Lipids, vol. 1867, no. 10, p. 159196, 2022.

[51] B. Wang et al., "Atherosclerosis-associated hepatic secretion of VLDL but not PCSK9 is dependent on cargo receptor protein Surf4," Journal of Lipid Research, vol. 62, 2021, Accessed: Mar. 17, 2024.